\documentclass[runningheads]{llncs}
\usepackage{graphicx}

\usepackage{tikz}
\usepackage{comment}
\usepackage{amsmath,amssymb} 
\usepackage{color}
\usepackage{multirow}
\usepackage{xspace}
\usepackage{xcolor}
\usepackage{cite}
\definecolor{citecolor}{HTML}{0071bc}
\definecolor{linkcolor}{HTML}{e802af}
\definecolor{urlcolor}{HTML}{00a600}
\definecolor{citecolor}{HTML}{0071bc}
\definecolor{linkcolor}{HTML}{e802af}
\definecolor{urlcolor}{HTML}{00a600}
\usepackage[pagebackref=true,breaklinks=true,colorlinks,citecolor=citecolor,linkcolor=linkcolor, urlcolor=urlcolor, bookmarks=false]{hyperref}

\usepackage[accsupp]{axessibility}  

\newcommand\dtdd{\textsc{DTD$^2$}\xspace}
\newcommand\dtd{\textsc{DTD}\xspace}

\begin{document}
\pagestyle{headings}
\mainmatter
\def\ECCVSubNumber{1}  

\title{How well does CLIP understand texture?} 

\titlerunning{How well does CLIP understand texture?}
%
\authorrunning{C. Wu, S. Maji}

\newcommand{\orcid}[1]{\,\href{https://orcid.org/#1}{\protect\includegraphics[width=8pt]{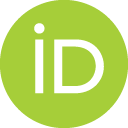}}}
\author{Chenyun Wu\orcid{0000-0001-6542-9279} \and
Subhransu Maji\orcid{0000-0002-3869-9334}}

\institute{University of Massachusetts Amherst \\ \email{\{chenyun, smaji\}@cs.umass.edu}}
\maketitle

\begin{abstract}
We investigate how well does CLIP understand texture in natural images described by natural language. To this end we analyze CLIP's ability to: (1) perform zero-shot learning on various texture and material classification datasets; (2) represent compositional properties of texture such as red dots or yellow stripes on the Describable Texture in Detail (\dtdd) dataset; and (3) aid fine-grained categorization of birds in photographs described by color and texture of their body parts. 
\end{abstract}

\section{Introduction}
Models with a joint understanding of language and vision such as CLIP~\cite{radford2021learning}, ALIGN~\cite{jia2021scaling}, and UNITER~\cite{chen2019uniter} have found their use in a number of applications ranging from guiding image generation and editing using language~\cite{ramesh2022hierarchical}, to open vocabulary object detection~\cite{gu2021open}, image segmentation~\cite{Luddecke_2022_CVPR,zhou2021denseclip}, and retrieval. These models are trained on massive collections of image and text pairs, and their remarkably good visual and language representations is reflected by their strong performance on many standard image understanding tasks. 

We focus on zero-shot learning capabilities of CLIP on the texture domains. Our motivation is two-fold. First, texture can be used to describe the appearance of a wide range of objects categories, especially in fine-grained domains. 
Second, there is a rich vocabulary to describe textures corresponding to color, pattern, structure, periodicity, stochasticity, and other properties. While prior work has evaluated zero-shot learning capabilities of CLIP on some texture datasets such as DTD~\cite{cimpoi14describing}, in this work we conduct a more detailed study in the context of texture understanding expanding in three directions. First, we incorporate a wider variety of datasets including FMD~\cite{Sharan-JoV-14}, KTH-TIPS~\cite{fritz2004kth} and KTH-TIPS2~\cite{mallikarjuna2006kth}. Second, we investigate CLIP's ability to handle compositional attributes of texture on \dtdd~\cite{wu2020dtd2}, which contains descriptions of textures in the DTD dataset. Third, we analyze how well CLIP recognizes texture attributes on real world images of birds described by the color and texture of body parts.

The study reveals that CLIP performs remarkably well on these tasks. Larger image encoders (e.g., ViT-L/14~\cite{dosovitskiy2020image}) are consistently better for zero-shot classification. Prompt tuning has a smaller impact on larger image encoders. CLIP also handles compositions well, and outperforms custom models trained on specific domains like DTD, especially in their ability to handle rare color and pattern combinations.
On the CUB dataset~\cite{WahCUB_200_2011}, the top 10 classification accuracy on zero-shot learning improves from $24.9\%$ to $56.6\%$ when texture attributes are added to the categories described by their scientific names. However, we also find that image encoders of CLIP have a significant foreground bias which can be problematic when referring to background regions and non-central objects.

\section{Background}

\paragraph{Zero-shot learning with CLIP.} Models such as CLIP, ALIGN, UNITER jointly learn an image encoder $\mathbf{\Theta}$ and a text encoder $\mathbf{\Phi}$ such as $\mathbf{\Theta(x)} \approx \mathbf{\Phi(y)}$ for image-text pairs ${(\mathbf{x},\mathbf{y})}$. CLIP uses bidirectional encoder representations using transformers (BERT~\cite{vaswani2017attention}) for text, and convolutional networks (e.g., ResNet~\cite{he2016resnet}) or transformer (ViT~\cite{dosovitskiy2020image}) encoders are used for images. 
CLIP was trained on a massive, curated dataset of 400 million image-text pairs, resulting in encoders that have good generalization abilities across visual recognition tasks.
To use CLIP for zero-shot learning, the description of a category $\textbf{y}$, often referred to as a prompt $\mathbf{p(y)}$, is encoded using the language model to obtain class prototypes $\mathbf{\Phi(p(y))}$. 
The encoded images are then classified based on the similarity to the class prototypes. 
The authors of CLIP showed that this strategy works remarkably well for a wide variety of datasets in computer vision. 
While the class label can be directly used as the prompt, i.e., $\mathbf{p(y) = y}$, better prototypes can be obtained by using a structure phrase involving the category as a prompt, e.g., ``a photo of a cat'' instead of ``cat". Prompts reflect the style of text accompanying images on the web and can significantly impact performance. 
While a large literature exists on designing prompts~\cite{liu2021pre,li2021prefix,lester2021power}, we explore a small space of hand-designed prompts on zero-shot texture recognition.

\paragraph{Texture datasets and tasks.} While many texture datasets exist in the literature, we focus on those that reflect describable properties of textures to benchmark CLIP's ability for zero-shot learning. This includes the Flickr material dataset (FMD) containing 10 material categories (e.g., wood, paper), Describable texture dataset (DTD) containing 47 describable attributes of texture (e.g., swirly, banded, zigzagged), KTH-TIPS and KTH-TIPS2a containing materials 10 and 11 materials taken under different lighting conditions. We also use the \dtdd dataset where images of DTD are annotated with descriptions of each image. \dtdd, unlike DTD contains multiple attributes that describe texture patterns in a compositional manner (e.g., red polka-dots or multicolored banded). We also explore the use of these attributes to describe species of birds in CUB Dataset~\cite{WahCUB_200_2011}. The dataset contains color and texture attributes of various body parts of the bird for each individual image. We collect these statistics at the entire set of images within a category to construct texture and color based descriptions of each category, which serve as a basis for zero-shot recognition. We compare the performance of the model against simply using the category name. CLIP has already seen many examples of each bird category associated with the images (possibly the entire CUB dataset is part of its training set). To simulate zero-shot learning on novel categories we compare against a baseline where we use the scientific names of the birds instead of their common ones. CLIP performs poorly in this setting, but by incorporating color and texture attributes the results improve significantly. 

\section{Experiments}
\subsection{Zero-shot texture classification}
\paragraph{Datasets and evaluation metrics.} We report the average per-image accuracy on all four datasets. For DTD we evaluate on the first split (``test1.txt") with 1180 images across 47 classes. For FMD we evaluated on the entire dataset (10 material classes with 100 images per class). For KTH-TIPS we evaluate on 810 images equally split across 10 classes with categories such as ``brown bread", ``sandpaper", ``cotton", etc. While for KTH-TIPS2a we use 4608 images across 11 classes, with roughly the same number of images per class. The datasets are publicly available and linked via the project's github repository \url{https://github.com/ChenyunWu/CLIP_Texture}. 

\paragraph{Results.} Table~\ref{tab:clip-texture-perf} shows the zero-shot classification accuracy of CLIP on DTD, FMD and KTH datasets. For this task we use the prompt ``a photo of a [c] pattern" for each category ``c" in the dataset. We observe that transformer variants (e.g., ViT-B/32 and ViT-L/14) are generally better than the ResNet (e.g., RN50 and RN101) counterparts.
The ViT-L/14@336 transformer trained on larger images (336$\times$336 vs. 224$\times$224) performs the best.
Table~\ref{tab:clip-prompt-perf} shows how the accuracy varies across prompts for two different image encoders. The best prompt varies across datasets but on average ``a photo of a [c] object" and ``a photo of a [c] pattern" performs best. Prompts have a larger impact on the performance of the smaller ViT-B/32 model compared to ViT-L/14 model indicated by the larger variance in performance across prompts. Table~\ref{tab:clip-texture-more} shows some additional results using the ViT-L/14@336 image encoder. The accuracy on FMD reaches 93.6\% using this encoder for the prompt ``a photo of [c] object", surprisingly outperforming current state-of-the-art ($\approx$ 85\%) based on bilinear representations and their variants~\cite{cimpoi2015deep,lin2015bilinear,lin2018second,gao2019global}. This could indicate potential overlap between FMD and the training set of CLIP. We are unable to verify this as the training dataset of CLIP is not publicly available, nor described in detail in the original paper.

\begin{table}[t]
\setlength{\tabcolsep}{5pt}
\small
    \centering
     \caption{\textbf{Performance of CLIP on zero-shot texture recognition.} Zero-shot accuracy for various image encoders using ``a photo of a [c] pattern" as the prompt.}
    
    \begin{tabular}{l|cccc|c}
    Model &	DTD	& FMD & KTH &	KTH2a & Average \\
    \hline
    RN50 &	40.7 &	83.4 &	49.1 &	62.8 & 59.0\\
    RN101 & 42.0 &	79.0 &	48.5 & 	51.3 & 55.2 \\ 
    ViT-B/32 &	41.1 &	83.8 &	58.4 &	59.5 & 60.7 \\
	ViT-B/16 &	44.7 &	87.9 &	57.4 &	61.1 & 62.8\\
	ViT-L/14 &	50.4 &	89.5 &	63.5 &	64.5 & 67.0 \\
	ViT-L/14@336 &	50.7 &	90.5 &	63.9	& 66.0 & 67.8 
    \end{tabular}
\label{tab:clip-texture-perf}
    \vspace{-2em}
\end{table}

\begin{table}
\setlength{\tabcolsep}{1pt}
\small
    \centering
        \caption{\textbf{Effect of prompt tuning.} Zero-shot accuracy for different prompts and image encoders. Prompt tuning has a larger impact on some datasets (e.g., KTH) and encoders (e.g., ViT-B/32). The best prompt varies across the datasets and encoders.}
    \label{tab:clip-prompt-perf}
    \begin{tabular}{l|cccc|cccc}
    & \multicolumn{4}{c|}{ViT-B/32} & \multicolumn{4}{c}{ViT-L/14} \\
    Prompt  	&	DTD	& FMD & KTH &	KTH2a &	DTD	& FMD & KTH &	KTH2a\\
    \hline
    [c]                        &	41.1 &	80.0 &	48.6 &	46.7  & 50.4 &  88.7 &	58.3 &	68.0 \\
    a photo of a [c]          &   43.1 &	79.9 &	50.4 &	49.9  &	52.3 &	89.0 &	61.0 &	69.4 \\
    a photo of a [c] background      &	43.1 &	79.9 &	50.4 &	49.9  &	50.4 &	89.3 &	59.3 &	69.8 \\
    a photo of a [c] object    &	42.3 &	83.2 &	56.3 &	59.7  &	53.0 &	92.3 &	59.6 &	70.0 \\
    a photo of a [c] pattern   &	41.1 &	83.8 &	58.4 &	59.5  &	50.4 &	89.5 &	63.5 &	64.5 \\
    \hline
    std. dev. & $\pm$1.0	& $\pm$2.0	& $\pm$4.3	& $\pm$6.0	& $\pm$1.3	& $\pm$1.5	& $\pm$2.0	& $\pm$2.3
    \end{tabular}

    \vspace{-2em}
\end{table}

\begin{table}
\setlength{\tabcolsep}{4pt}
\centering
\small
 \caption{\textbf{Additional results using ViT-L/14@336 image encoder.} Zero-shot accuracy is shown on four texture datasets.}
    \begin{tabular}{l|c|cccc|c}
    Prompt & Model &	DTD	& FMD & KTH &	KTH2a & Average \\
    \hline
	a photo of a [c] object & ViT-L/14@336 &	53.3 &	93.6 &	59.4	& 69.5  & 69.0\\
	a photo of a [c] pattern & ViT-L/14@336 &	50.7 &	90.5 &	63.9	& 66.0  & 67.8
    \end{tabular}
    \label{tab:clip-texture-more}
\end{table}

\subsection{Performance on Describable Texture in Detail Dataset}


\begin{table*}[t]
\setlength{\tabcolsep}{8pt}
\centering
  \caption{\textbf{Retrieval performance of DTML and CLIP on \dtdd.} Various performance metrics on image and phrase retrieval are shown for CLIP and DTML.}
    \begin{tabular}{c|c|*{6}{c}}
    Task & \multicolumn{1}{c|}{Model} & \multicolumn{1}{c}{MAP} & \multicolumn{1}{c}{MRR} & P@5 & \multicolumn{1}{c}{P@20} & \multicolumn{1}{c}{R@5} & \multicolumn{1}{c}{R@20} \\
    \hline
    \multirow{2}[2]{*}{Phrase retrieval} 
          & {DTML} & \textbf{31.6} & \textbf{72.5} & \textbf{40.6} & \textbf{22.9} & \textbf{20.2} & \textbf{44.5}  \\
          
           & CLIP & 12.2 & 40.0 & 17.6 & 11.4 & 8.4 & 21.5\\
    
    \hline
    \multirow{2}[2]{*}{Image retrieval} 
     & DTML & \textbf{13.5} & 31.1 & 16.5 & \textbf{14.5} & 5.2  & \textbf{17.3} \\
     & CLIP & 12.7 & \textbf{32.1} & \textbf{16.9} & 13.2 & \textbf{6.1} & 17.3\\
    \end{tabular}
  \label{tab:dtd2}
\end{table*}



\begin{figure}
\centering
   \includegraphics[width=\linewidth]{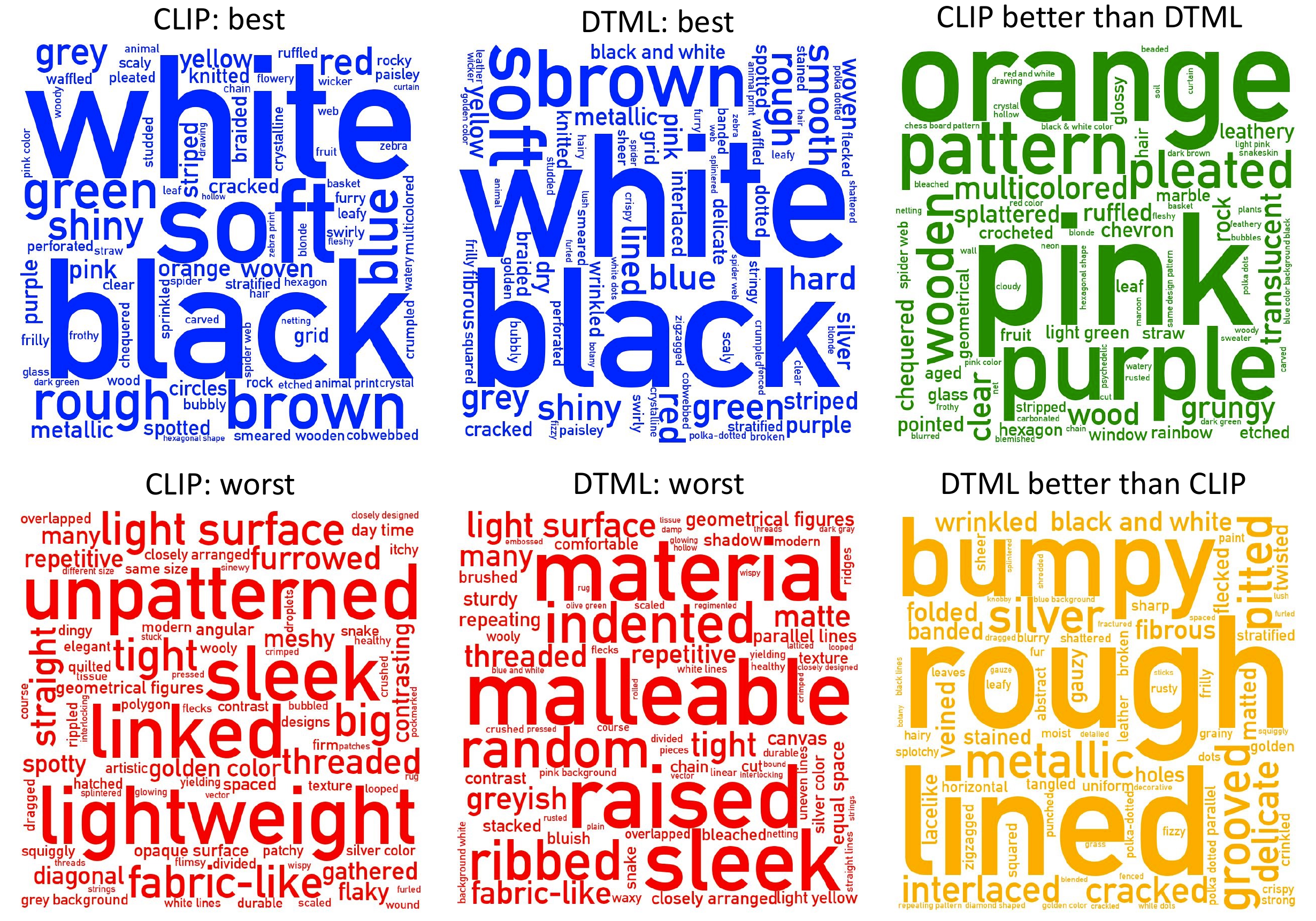}
   \vspace{-1em}
   \caption{\textbf{Best and worst performing attributes on \dtdd for CLIP and DTML.} Each cloud represents top and bottom 80 attributes based on average precision on the retrieval task for CLIP (left) and DTML (right). On the right 80 attributes with the highest difference in performance between the two models. Font sizes of attributes are proportional to their frequency in \dtdd.}
   \label{fig:dtd2_cloud}
\end{figure}

We first compare CLIP with DTML on phrase and image retrieval on \dtdd. DTML is the metric-learning baseline presented in~\cite{wu2020dtd2}, which trains an off-the-shelf BERT text encoder and a ResNet101 image encoder using a triplet-based metric learning loss on the \dtdd dataset. A linear layer on top of BERT and all the layers of ResNet101 are trained (or fine-tuned). For phrase retrieval we rank the 655 frequent phrases in \dtdd according to their distances to the query image, while for image retrieval we rank the images in the test set based on distances to the query text. For CLIP we use the prompt ``an image of [c] texture" where ``c" is the category. All results for CLIP are using the ViT-B/32 image encoder.

Table~\ref{tab:dtd2} shows the retrieval performance of CLIP and DTML on \dtdd. 
CLIP obtains similar performance on image retrieval but is worse on phrase retrieval compared to DTML.
Figure~\ref{fig:dtd2_cloud} shows the best and worst attributes for each model. 
We calculate the image retrieval average precision (AP) for each phrase and plot top and bottom 80 phrases. 
We also visualize phrases with the largest difference of AP between the two models. 
The two models are both good at phrases that describe common colors and patterns, but their worst performing phrases are different.
CLIP is better than DTML on rare colors such as ``orange", ``pink", ``purple". CLIP is also better on attributes related to materials or certain types of objects (e.g., ``wood", ``marble", ``glass") with are relatively rare. However, CLIP performs worse used to describe patterns and textures frequent in \dtd (e.g., ``rough", ``lined", ``grooved").

\begin{table*}[t]
\setlength{\tabcolsep}{4pt}
\centering
\caption{\textbf{R-precision of image retrieval on \dtdd.} CLIP understands compositional attributes despite not trained on this dataset. For example, on the ``two-colors" the performance is significantly better. It also exhibits a significant foreground bias as indicated by the lower performance on the ``background" task.}
    \begin{tabular}{c|cccc}
        \textbf{Model} & \textbf{~Foreground~} & \textbf{Background~} & \textbf{Color+Pattern~} & \textbf{Two-colors}\\
        \hline
        DTML & \textbf{46.5$\pm$20.6} & 52.0$\pm$6.3 & 41.7$\pm$22.8 & 27.4$\pm$15.1\\
        CLIP & 38.0$\pm$14.9 &	\textbf{60.2$\pm$5.5} &	\textbf{45.2$\pm$23.5} &		\textbf{55.2$\pm$16.2} \\
        \hline
        Chance & 50.0 & 50.0 & 7.4 & 5.5\\
\end{tabular}
\label{tab:clip:comp}
\end{table*}

\paragraph{Attributes as prompts.}
\dtdd contains multiple attributes for each image which could be incorporated into the prompt design for each category. We include the 20 most frequent attributes for each category in the prompt as ``an image of [$p_1$, $p_2$,\ldots, $p_{20}$] texture", where $p_i$ is the $i^{th}$ most-frequent phrase. 
For example, the ``gauzy" category is described as ``an image of gauzy, sheer, transparent, light, thin, white, translucent, soft, see through, delicate, netted, meshy, airy, silky, fabric, see-through, folded, wavy, curtains, cloth texture."
This improves the accuracy to {54.8\%} from 41.1\% when only including the category as prompts with the ViT-B/32 encoder on DTD.

\paragraph{Synthetic Textures.} We conduct the compositionality modeling analysis on synthetic texture images the same as described in Section 5.3 of~\cite{wu2020dtd2}. 
Given a query phrase such as ``blue and red", the task is rank the positive and hard negative images (which are ``blue" or ``red" but not both). The R-precision for the retrieval task is listed in Table~\ref{tab:clip:comp}. CLIP achieves a significant improvement on ``Two-colors" and a slight improvement on ``Color+Pattern" over the DTML model. 
The larger training set of CLIP allows better generalization to rare or novel combinations in \dtdd. 
We also see a slight improvement for CLIP on ``Background" compared against DTML but it performs lower than random guesses on ``Foreground". This suggests that CLIP 
likely has a foreground bias. We investigate this aspect further on CUB Dataset.

\subsection{Performance on Caltech-UCSD Birds Dataset}

\begin{figure*}
\centering
   \includegraphics[width=\linewidth]{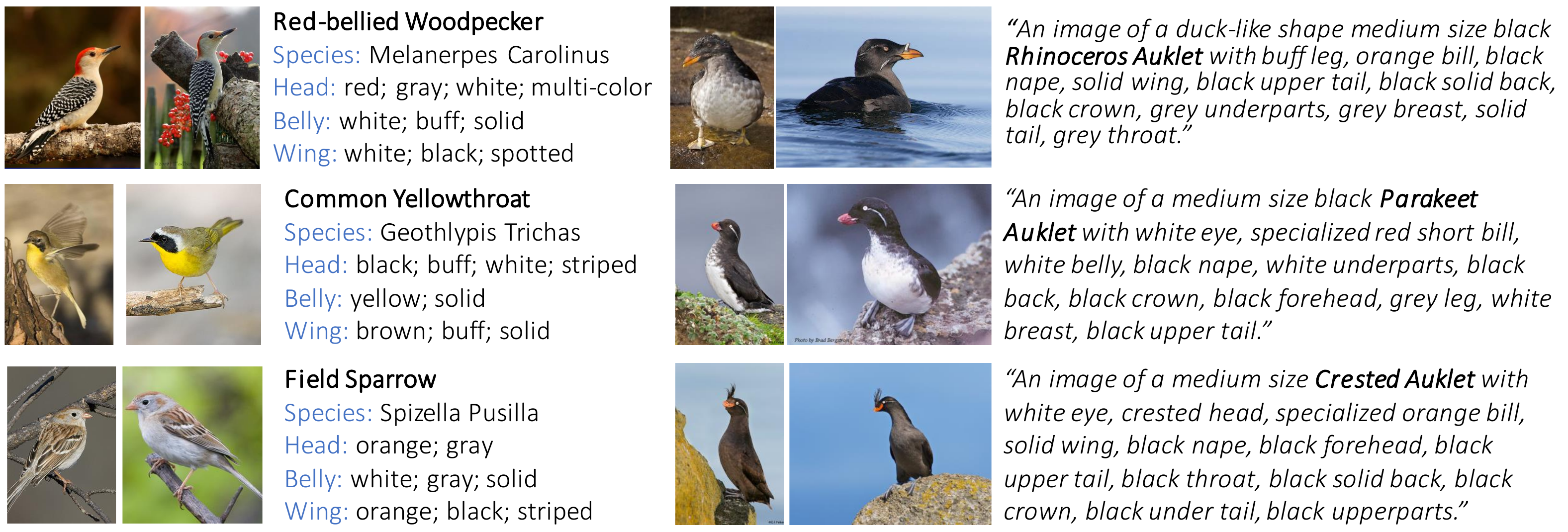}
   \vspace{-2em}
   \caption{\textbf{Examples of birds and their attributes.} 
   On the left are some bird species with the CUB annotations indicating color and texture of body parts. On the right are automatically generated prompts based on these attributes for zero-shot learning.
   }
   \label{fig:classify_eg}
\end{figure*}

\begin{table*}[t]
\centering
\setlength{\tabcolsep}{4pt}
  \caption{\textbf{Phrase and image retrieval on CUB.} We experiment with 17 attributes that are included in both CUB and \dtdd.}
    \begin{tabular}{c|c|*{6}{c}}
    Task & \multicolumn{1}{c|}{Model} & \multicolumn{1}{c}{MAP} & \multicolumn{1}{c}{MRR} & P@5 & \multicolumn{1}{c}{P@20} & \multicolumn{1}{c}{R@5} & \multicolumn{1}{c}{R@20} \\
    \hline
    \multirow{2}[2]{*}{Phrase retrieval} 
          & DTML & 52.6 & 68.6 & \textbf{46.4} & - & \textbf{45.8} & - \\
           & CLIP & \textbf{54.1} & \textbf{75.9} & 43.2 & - & 43.4 & -  \\
    
    \hline
    \multirow{2}[2]{*}{Image retrieval} 
     & DTML & 35.3 & 53.7 & 44.7 & 43.8 & 0.2 & 0.7 \\
     & {CLIP}  & \textbf{50.1} & \textbf{91.7} & \textbf{72.9} & \textbf{71.8} & \textbf{0.5} & \textbf{1.6}\\
    \end{tabular}
\label{tab:clip_cub}
\end{table*}

\begin{figure*}
\centering
   \includegraphics[width=\linewidth]{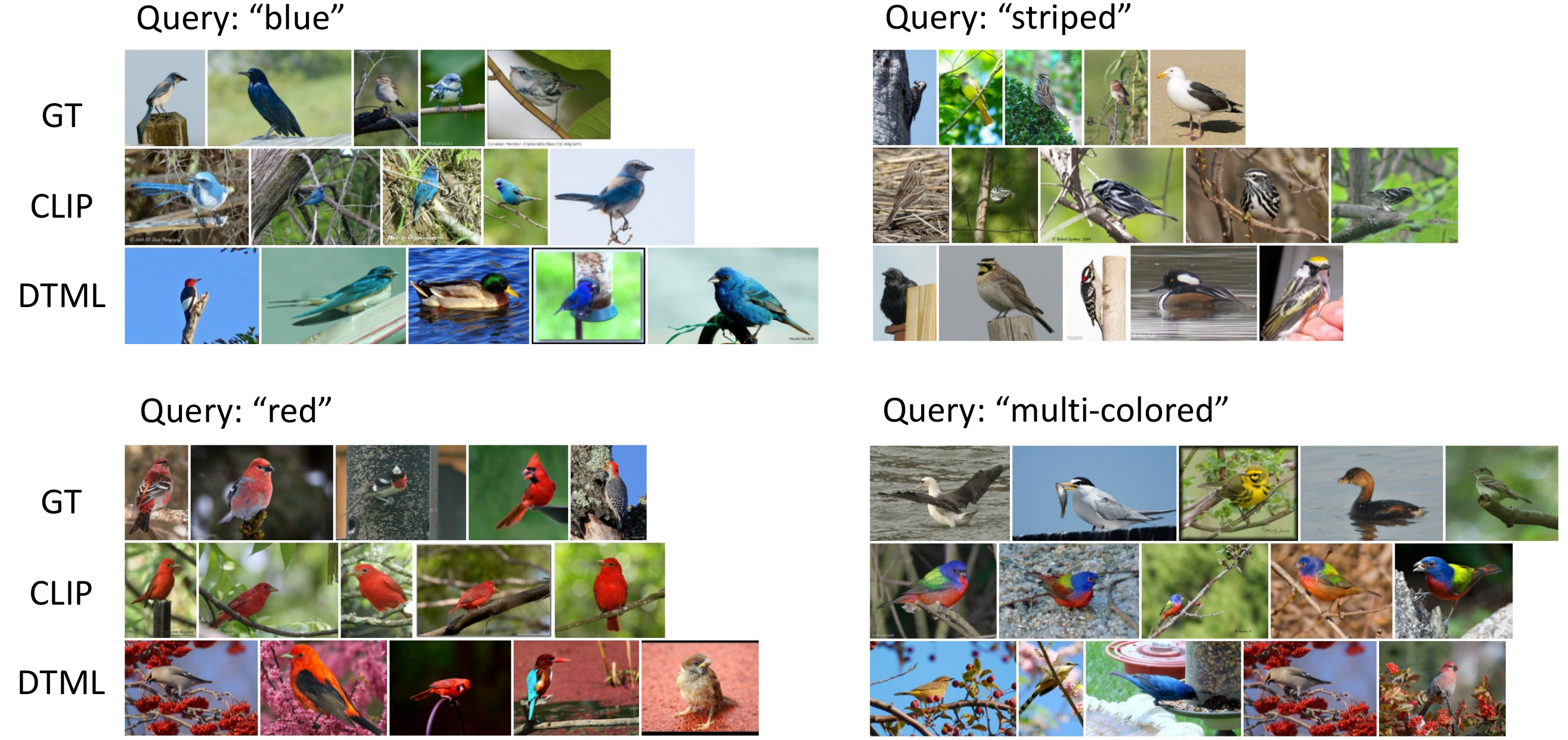}
   \vspace{-2em}
   \caption{\textbf{Image retrieval on CUB.} For each query we show 5 ground-truth and top images retrieved using CLIP and DTML. CLIP primarily associates attributes with the foreground ---  e.g., the query ``blue" is only associated with blue birds for CLIP, while the DTML retrieves both blue birds and the background water.}
   \label{fig:cub_eg}
\end{figure*}


Once again, we compare the two models, DTML and CLIP, on the CUB dataset \cite{WahCUB_200_2011}. 
Images in this dataset differ from the types of images in DTD which poses a significant domain shift for DTML. 
For evaluation we select 17 attributes that both occur in \dtdd and CUB dataset. 
Images with an attribute on any part, i.e., ``striped" on wing, upper-parts, or back, are counted as positives for the attribute, i.e., ``striped".

Table~\ref{tab:clip_cub} shows the retrieval performance of DTML and CLIP. 
CLIP performs better than DTML on image retrieval and they perform similarly on phrase retrieval.
Figure~\ref{fig:cub_eg} shows example retrieved images. CLIP focuses on the foreground while DTML recognizes attributes from the background as well. For example, CLIP retrieves ``blue" birds, while DTML retrieves images with a ``blue" background such as water.
DTML retrievals are from different categories, but CLIP tends to return images of the same category, which implies that CLIP image features are highly related to categories such that images of the same category are close to each other in the embedding space.

\begin{table*}
    \centering
    \caption{\textbf{Zero-shot classification top-k accuracy on CUB test set using CLIP with different levels of category names and attributes.} In the brackets we show the number of categories for each level, .e.g., there are 115 different genus names.}
    \begin{tabular}{c|cc|cc|cc|cc|cc|c}
         \textbf{category} & \multicolumn{2}{c|}{\textbf{name}(200)}  & \multicolumn{2}{c|}{\textbf{species}(200)} & \multicolumn{2}{c|}{\textbf{genus}(115)} & \multicolumn{2}{c|}{\textbf{family}(39)} & \multicolumn{2}{c|}{\textbf{order}(12)} & \textbf{``bird''}(1)\\
         \hline
         \textbf{\#attribute} & 0 & 15 & 0 & 15& 0 & 15& 0 & 15& 0 & 15&15\\
         \hline
         {top-1} & 51.8 & 50.2 & 6.6 & 14.3 & 15.4 & 14.4 & 10.8 & 12.9 & 6.6 & 12.5 & 12.1 \\
         {top-5} & 82.7 & 81.6 & 19.2 & 41.0 & 18.8 & 40.5 & 11.5 & 36.0 & 6.7 & 35.4 & 35.8 \\
         {top-10} & 91.0 & 91.3 & 24.9 & 56.6 & 20.8 & 53.3 & 17.3 & 51.9 & 11.3 & 51.5 & 52.3\\
    \end{tabular}
    \label{tab:cub_classify}
\end{table*}

\paragraph{Zero-Shot Classification with Attributes.} To construct a list of attributes for each category, for each attribute we estimate the ratio of images that contain the attribute within a category to the images that contain that attribute across all categories.
Top $k$ such attributes are added to prompts for the category in the format ``an image of a [$P$] [$C$] with [$P_1$] [$N_1$], [$P_2$] [$N_2$], ..."
where $C$ is the category label, $P$ are attributes of the entire bird, $N_i$ are body parts (such as ``belly", ``tail") and $P_i$ are attributes for the part $N_i$. 
Figure~\ref{fig:classify_eg} shows three examples of constructed descriptions. 
The attributes reflect subtle differences among the three similar species, e.g., ``Rhinoceros Auklet" is more ``duck-like" with ``buff leg", ``Parakeet Auklet" has ``white eye" and ``white belly", ``Crested Auklet" has ``crested head" and ``black nape".
We also use different choices for the category label, varying them from the common name to scientific names at different levels of the biological taxonomy. The common name is often used to describe the category and be associated with the corresponding images on the Internet. Scientific names at various levels in the taxonomy are less likely to have been observed by the language models.

Classification results are shown in Table~\ref{tab:cub_classify}.
We report the top-k accuracy for the 200-way classification.
When no attribute is added, we construct the description simply as “an image of a [$C$]” where [$C$] can be the common name, species, genus, family, or order.
In such cases species from the same genus, family or order have the same category embedding resulting in ties.
There is a significant performance drop when using scientific names compared to the common names. CLIP has learned plenty about the common names during training as it exploits images available on the web but fails when using species names which are relatively rare. Adding attributes to the descriptions improves performance by a large margin in this case, especially on the top 5/10 accuracy. With the help of attributes, we achieve similar performance when we replace the scientific names with the generic category description ``bird”. 
\section{Conclusion and Limitations}\label{sec:conclusion}
We analyze how well CLIP recognizes describable properties of texture in natural images. 
Remarkably, CLIP achieves strong zero-shot performance for texture classification, outperforming strong baselines on texture image and phrase retrieval on \dtdd.
Texture understanding is also effective on natural images of birds for attributes that describe the color and texture of body parts. This brings up the exciting possibility of applying similar models on other fine-grained domains such as fashion, Fungi, and Butterflies.
At the same time, we observe a foreground bias in the model, which might not be desirable when referring to attributes of the background or non-central object in the image. 
We hope this contributes to a better understanding of performance and biases of CLIP for various applications.

\clearpage
{
\bibliographystyle{ieee_fullname}
\bibliography{egbib}
}

\end{document}